\documentclass[11]{article}
\usepackage{preamble}

\usepackage[utf8]{inputenc}

\title{Neural Network Verification with PyRAT}
\author{Augustin Lemesle, Julien Lehmann, Tristan Le Gall}

\newcommand{\diff}[1]{\textcolor{blue}{#1}}

\newcommand{\pyrat}{\textsc{PyRAT}\xspace}
\newcommand{\tool}[1]{\textsc{#1}}
\newcommand{\wrt}{\textit{w.r.t.}\;}
\newcommand{\ie}{\textit{i.e.}\;}
\newcommand{\eg}{\textit{e.g.}\;}
\newcommand{\RR}{\mathbb{R}}
\newcommand{\zono}{\emph{zonotope}\xspace}
\newcommand{\zonos}{\emph{zonotopes}\xspace}
\newcommand{\conz}{\emph{constrained zonotope}\xspace}
\newcommand{\hybz}{\emph{hybrid zonotope}\xspace}
\newcommand{\conzs}{\emph{constrained zonotopes}\xspace}
\newcommand{\hybzs}{\emph{hybrid zonotopes}\xspace}
\newcommand{\bab}{\text{branch and bound}\xspace}
\newcommand{\Bab}{\text{Branch and bound}\xspace}
\newcommand{\relu}{\text{ReLU}\xspace}
\newcommand{\nn}{neural network\xspace}

\newcommand{\vnncomp}{VNN-Comp\xspace}
\newcommand{\vnnlib}{VNN-LIB\xspace}
\newcommand{\safeprop}{\textit{safety property}\xspace}
\newcommand{\Alpha}{\boldsymbol{\alpha}}
\newcommand{\mzono}[2]{\langle #1,\:#2\rangle}

\newcommand{\abcrown}{\textsc{$\alpha,\beta-$CROWN}\xspace}

\newcommand{\bnb}{BaB\xspace}

\newcommand{\cex}{counter\-example\xspace}
\newcommand{\cexs}{counter\-examples\xspace}

\DeclareMathOperator*{\argmax}{arg\,max}

\begin{document}

\maketitle

\begin{abstract}
    As AI systems are becoming more and more popular and used in various critical domains (health, transport, energy, ...), the need to provide guarantees and trust of their safety is undeniable. To this end, we present \pyrat, a tool based on abstract interpretation to verify the safety and the robustness of neural networks. In this paper, we describe the different abstractions used by \pyrat to find the reachable states of a neural network starting from its input as well as the main features of the tool to provide fast and accurate analysis of neural networks. \pyrat has already been used in several collaborations to ensure safety guarantees, with its second place at the \vnncomp 2024 showcasing its performance.
\end{abstract}

\section{Introduction}
There is no doubt that Artificial Intelligence (AI) has taken over an important part of our lives and its recent popularisation with large language models has anchored this change even more in our current landscape. The use of AI is becoming more and more widespread reaching new sectors such as health, aeronautics, energy, etc., where it can bring tremendous benefits but could also cause environmental, economic, or human damage, in critical or high-risk systems. In fact, numerous issues are still being uncovered around the use of AI, ranging from its lack of robustness in the face of adversarial attacks \cite{advex1, advex2}, to the confidentiality and privacy of the data used, the fairness of the decisions, etc.

Faced with these threats and the exponential growth of its use, regulations have started to emerge with the European AI Act. 
Not waiting for regulations, tools have already been developed to respond to and mitigate these threats by providing various guarantees from the data collection phase to AI training and validation of the AI. Our interest here lies in this last phase, the formal validation of an AI system, and more specifically a neural network, to allow its use in a high-risk system.

Neural networks are a subset of AI models and are also one of the most popular and widely used architectures. They include a large variety of architectures (DNN, CNN, GNN, Residual, Transformers, etc.) applicable to various use cases such as image classification, regression tasks, detection, control \& command, etc. Similarly, to a classical software, these neural networks, when implemented in critical systems, should be tested and validated in order to provide strong guarantees regarding their safety. 
While there are many approaches, such as empirically testing the robustness to small input variations, adversarial attacks or even metamorphic transformations \cite{aimos}, we focus here on using formal methods to provide strong mathematical guarantees on the safety of a \nn. 

To this end, \pyrat, which stands for Python Reachability Assessement Tool, has been under development at CEA since 2019, in the hope of simplifying the validation process of \nn. It provides an simple interface and an easy step to formally prove the safety of \nn, which can often be seen as an arduous task. To easily interface with most of the neural network training frameworks (Tensorflow, PyTorch, etc.) \pyrat is written in Python and handles several standard AI formats (ONNX, NNet, Keras, etc.). \pyrat also provides several possibilities and interfaces to evaluate the safety of a network through standard safety property files (\vnnlib or similar format) or through several Python APIs.

\pyrat has been time-tested in several national and international projects as well as in multiple collaboration with industrial partners. It has also participated in the international neural network verification competition, the \vnncomp, in 2023 and 2024 \cite{vnn2023} achieving third and second place respectively.

\section{Related tools}
Even before the widespread use of machine learning, classical programs had to be formally verified and/or certified to avoid any erroneous behavior when used in critical systems. Thus, many verification techniques used for AI were initially developed to increase trust in critical systems, including \emph{model-checking}~\cite{modelcheckingClarkeEmerson,modelcheckingClarkeEmersonSifakis}, which represents the program as a state transition system and checks whether the model built satisfies a formal property, often using Satisfiability Modulo Theory (SMT), or \emph{abstract interpretation}~\cite{cousot:article,cousot:book}, which abstracts the semantics of a program to compute an approximation of what the program does (\eg an over-approximation of the reachability set).

As the need arose, these methods were later adapted to the verification of neural networks with specific modifications, \eg no loops, no memory to allocate, but instead linear and non-linear operation on huge matrices or tensors. The first tools developed for this purpose were SMT-based methods~\cite{pulina2012smt, SMT4vnn, reluplex}, which rely on SMT formulations and solvers to compute the feasibility of an input-output property on small neural networks.
Instead of using SMT, some tools transformed the verification task into a constrained optimisation problem in the Mixed Integer Programming (MIP) setting \cite{MILP4vnn, dutta2017milp}. Both of these techniques are known to be computationally expensive.

To avoid these costly computations, methods based on abstract interpretation have been developed~\cite{reluval, ai2}. In general, tools using these methods represent the inputs they want to verify as a set which is then passed along the \nn and it yields a set that over-approximates the output which is used to evaluate the property. The formalism used to represent the set along the \nn is called an abstract domain. The accuracy of an analysis by abstract interpretation depends heavily on the abstract domain chosen. If the result of the analysis is too imprecise, one way to improve the precision is to combine several abstract domains~\cite{cousot:book,libra}. 

We briefly present a non-exhaustive list of tools aimed at the formal verification of neural networks with abstract interpretation and we refer to \citet{urban2021reviewformalmethodsapplied} for a more comprehensive survey. 
Many of these tools, including \pyrat, accept the Open Neural Network Exchange (ONNX) format to describe the neural network and the \vnnlib format (based on SMT-LIB) to describe the properties to be verified and have participated to at least one instance of the annual \vnncomp. They are marked with (*).

\begin{itemize}
    \item (*)\tool{alpha-beta-CROWN}~\cite{crown,alphacrown,betacrown} uses the CROWN-domain which is mathematically equivalent to the polytope domain with a tailored abstractions for \relu and other non-linear functions, and \bab verification strategy. This tool won the yearly \vnncomp from 2021 to 2024.
    \item (*)\tool{MN-BaB}~\cite{mn-bab} is a dual solver built on the DeepPoly domain~\cite{deeppoly} leveraging multi-neural relaxation of the \relu along with \bab verification strategy.
    \item (*)\tool{NNV}~\cite{nnv1, nnv2} and (*)\tool{nnenum}~\cite{nnenum1, nnenum2} use the star set domain that uses linear programming to optimize ReLU abstractions for the Zonotopes domain.
    \item \tool{Libra}~\cite{libra} also uses abstract interpretation, although for checking fairness properties (unlike \pyrat and other tools that analyse reachability and safety properties). Like \pyrat, \tool{Libra} can combine several abstract domains to increase the precision of its analysis. Its domains include intervals, as well as domains from the \tool{Apron}~\cite{Apron} library or domains reimplemented from \tool{Symbolic}~\cite{symbolic}. \tool{Deeppoly}~\cite{deeppoly}, and \tool{Neurify}~\cite{neurify}.
    \item \tool{Saver}~\cite{saver-vmcai24,saver-sas19} verifies properties on support vector machines (SVMs), focusing on robustness or vulnerability properties of classifiers. Its domains include Intervals and Zonotopes.
\end{itemize}

\pyrat also relies on abstract interpretation techniques to verify the safety of neural networks. It implements several abstract domains such as intervals, zonotopes, zonotopes paired with inequality constraints or a DeepPoly reimplementation. These domains can be combined and optimised by \pyrat with multiple strategies such as \bab.

\section{Key Notions}
\subsection{Neural Networks}
Neural networks are functions $N: \RR^n \to \RR^m$ which are generally defined in terms of layered architectures alternating between linear and non-linear operations. 

\begin{figure}[ht!]
    \centering
    \includegraphics[height=0.5in]{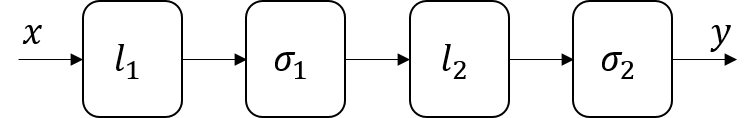}
	\caption{Feedforward neural network.}
	\label{fig:feedforward}
\end{figure}

We represent a neural network as a directed graph of layers resulting in a composition of the operations corresponding to each layer. For example, a simple feedforward neural network with an activation function between each layers without any loop or branching in the network as shown in Figure \ref{fig:feedforward}.
~

It can also be defined as the composition of each layer of the network with $x \in \RR^n$ the input and $y \in \RR^m$ the output of the network:
$$y = N(x) = \sigma_2(l_2(\sigma_1(l_1(x)))).$$

\begin{figure}[h!]
    \centering
    \includegraphics[height=0.7in]{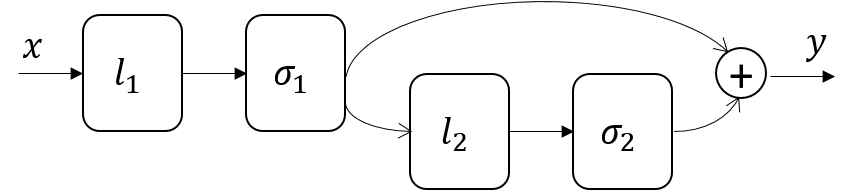}
	\caption{Residual neural network.}
	\label{fig:residual}
\end{figure}

Typically, the activation functions $\sigma_1$ and $\sigma_2$ can be the Rectified Linear Unit (\relu), the Sigmoid or the Tanh functions which are defined below. While the layers $l_1$ and $l_2$ can be matrix multiplication or convolution layers.
\[
\begin{array}{lcll}
\relu(x) &= &\max (x, 0) &\\[6pt]
\text{Sigmoid}(x) &= &\dfrac{1}{1 + e^{-x}} &\\[13pt]
\text{Tanh}(x) &= &\dfrac{e^x - e^{-x}}{e^x + e^{-x}} &= 2 \times \text{Sigmoid}(2x)-1 \\[8pt]
\end{array}
\]
More complex architectures can include some residual connections between layers to model more complex behaviors, as represented in Figure \ref{fig:residual}.

The representation in Figure \ref{fig:residual} is equal to the following formula:
$$y = N(x) = \sigma_2(l_2(\sigma_1(l_1(x)))) + \sigma_1(l_1(x)).$$

Overall, we classify the layers of a neural network into three categories which will lead to different treatment in \pyrat:
\begin{itemize}
    \item Linear layers such as affine or convolutional layers.
    \item Non-linear layers such as the various activation functions but also layers implementing multiplication or division between different layers (Mul, Div, Pow, ...).
    \item The layers that do not directly influence the values of the data but may be only modify their shape such as concatenation, flattening or reshaping layers.
\end{itemize}

\subsection{Reachability Analysis}

Let $N: \mathbb{R}^n \to \mathbb{R}^m$ be a neural network.
We define the notion of \safeprop by a set of constraints on the inputs and outputs of $N$. Through set inclusion, we express the satisfiability of this \safeprop: given the input set $X \subseteq \mathbb{R}^n$ and the output set $Y \subseteq \mathbb{R}^m$ defined by the \safeprop, the \safeprop is satisfied if and only if $N(X) \subseteq Y$. 

Ideally, we would like to define a \safeprop such that for all inputs $x \in \RR^n$ of a given neural network $N$, the output $y$ is correctly classified by $N$. However, in practice, we cannot define the correct classification for all inputs since this is precisely the task we are trying to solve using a neural network. Therefore, we can only reduce the scope of what we aim to verify by defining more general properties.

The first type of \safeprop here consists of properties on the behavior of the models. For example, given a certain input space $X$, we know that the output will always be $y$. These properties are well suited for control-command neural networks such as ACAS-Xu \cite{acasxuintrod} where the data are structured around some physical values and their semantic is clear. For ACAS-Xu, \citet{katz2017reluplex} defined 10 such safety properties on the neural networks used. For example, property 1 is expressed by:
\begin{itemize}
    \item Description: If the intruder is distant and is significantly slower than the ownship, the score of a Clear-of-Conflict advisory will always be below a certain fixed threshold.
    \item Input constraints: $\rho \ge 55947.691, v_{own} \ge 1145, v_{int} \le 60$
    \item Desired output property: the score for Clear-of-Conflict is at most 1500.
\end{itemize}

More generally for the ACAS-Xu, as we possess a fully defined specifications through a look up table (LUT), the \safeprop can be formalised like in \citet{damour2021} by saying that the decision of the neural network should be contained in the decision of the LUT:
$$\forall x \subseteq \RR^5, N(x) \subseteq LUT(x)$$

On the other hand, with unstructured data such as images or sound, it is harder to define a semantically meaningful set as the input space that we want to verify, \eg  "there is a pedestrian on the image". Defining what a pedestrian is on an image is an open question and approaches such as using simulators or image generation may answer this in the future \cite{girard2020camus}. For now, we focus on a more local approach to the safety on such data. 
For example, considering a classification task, we can define a local robustness property as follows: 

\underline{Local robustness:} Given an input $x_0 \in \mathbb{R}^n$ and a modification size $\epsilon > 0$, we want to prove or invalidate the following property
$$
\forall x' \in B_{\|.\|_\infty}(x_0, \epsilon), \,\, \argmax_{i} N(x_0)_i = \argmax_i N(x')_i
$$
where $B_{\|.\|_\infty}(x_0, \epsilon) := \{z \,|\, \|x_0 - z\|_\infty \leq \epsilon \}$. 

In this example, we have a set $X := B_{\|.\|_\infty}(x_0, \epsilon)$ and a set $Y := \{ y'  \, | \, \argmax_{i} N(x_0)_i = \argmax_i y'_i \}$, and the local robustness property holds if $N(X) \subseteq Y$. The local robustness property can be extended to a regression task, \ie without the argmax by adding a threshold on the distance between the two outputs.

Computing $N(X)$ for any given input set $X \subseteq \RR^n$ is called a \textit{reachability analysis}, and its exact computation is generally hard. For example, when the activation function of $N$ are only \relu, then computing exactly $N(X)$ is NP-complete~\cite{reluplex}.
Therefore, \pyrat uses \textit{abstract interpretation} to compute an over-approximation of the reachability set $N(X)$.

\subsection{Abstract Interpretation}

Abstract Interpretation~\cite{cousot:article,cousot:book} is a theoretical framework that can be applied to the verification of computer programs and neural networks. Its principle is to abstract the semantics of a program and thus to compute an over-approximation of its reachability set by the following method:
\begin{itemize}
    \item Choose an abstract domain $\Lambda^\sharp$, and two operators $\alpha: \mathcal{P}(\RR^d) \to \Lambda^\sharp $ and $\gamma: \Lambda^\sharp \to \mathcal{P}(\RR^d)$
    forming a \textit{Galois connection}~\cite{cousot:book}
    \item Determine, for each layer $N_j$, a safe abstraction $N_j^\sharp: \Lambda^\sharp \to \Lambda^\sharp$ such that $ \alpha \circ N^j \circ \gamma \sqsubseteq N_j^\sharp: $
    \item Compute $ N^\sharp(\alpha(X))$ by applying successively the abstraction of each layer to either $\alpha(X)$ (if it is the first layer) or to the outputs of (the abstraction of) the previous layer
    \item The \textit{Galois connection} ensures that $N(X) \subseteq \gamma(N^\sharp(\alpha(X)))$
\end{itemize}

The classical Abstract Interpretation framework assumes that $\Lambda^\sharp$ is a complete lattice~\cite{cousot:book} (e.g. the Interval lattice); however, the approach works even when the lattice $\Lambda^\sharp$ is not complete (e.g. convex polyhedra) and even when $\Lambda^\sharp$ is not a lattice (e.g. zonotopes), as long as we can define safe abstraction of each operator. 

Therefore, this approach is parameterised by the abstract lattice $\Lambda^\sharp$ and how we abstract each layer, especially when the layer implements a non-linear function. We will explain these choices in \pyrat in the next sections. 

Through the \textit{Galois connection}, abstract interpretation ensures that the resulting abstraction of the neural network correctly over-approximates its true reachable sets. As such, \pyrat is said to be \textbf{correct}, \ie \pyrat will only say that a safety property is verified or falsified if it is mathematically sure of the result, otherwise if the abstraction is not tight enough the result will be unknown.

\subsection{Soundness w.r.t. real arithmetic} 
In this paper, we consider the mathematical definition of the \nn operations, i.e. a semantics based on real numbers. However, computers use floating-point arithmetic, which introduces, usually small, rounding errors. In order to be sound with respect to the  real arithmetic, \pyrat captures the rounding error by choosing the most pessimistic rounding for each operation. This means that if the result of a real number computation is $x$, then \pyrat will compute a range $[\underline{x},\overline{x}]$ where $\underline{x}$ and $\overline{x}$ are two floating-point numbers such as $\underline{x} \leq x \leq \overline{x}$. We refer to \citet{soundZono} (resp. \cite{soundPoly}) for a presentation of an abstract domain sound w.r.t real arithmetic based on Zonotope (resp. Polyhedra). 
Note that the sound mode of \pyrat is only available in CPU and can be disabled for faster analysis. 

\section{Abstract domains \& abstract transformers}

In this section, we present several domains to abstract  $\mathcal{P}(\RR^d)$ and the operation commonly found in neural networks.

\subsection{Boxes and Interval Arithmetic}

One of the easiest ways to abstract any set $ X \in \mathcal{P}(\RR^d)$ is to abstract each dimension individually (we consider the projection of $X$ on dimension $i$: $X_i = \mathrm{proj}_i(X)$). We thus obtain a \emph{box} $\alpha(X) = (x_1, \dots, x_d)$ where, for each dimension $i$, $x_i= [\underline{x}_i, \overline{x}_i]$ is an interval with $\underline{x}_i = \min X_i$ and $\overline{x}_i = \max X_i$. This is the first abstract domain implemented in \pyrat.

For each dimension, we can then do interval arithmetic to compute the result of different operations\footnote{in this section, we only present intervals with finite (real number) bounds and refer to \citet{interval_arithmetic,dawood2011theories} for the extension to intervals with floating-point or infinite bounds} \cite{moore1966interval}: 
\[
\begin{array}{|l|c|l|}
\hline
\text{addition} & [a,b] + [c,d] & [a+c,b+d]\\
\hline
\text{opposite} & -\, [a,b] & [-b,-a]\\
\hline
\rule{0pt}{3.8ex}
\text{scalar multiplication} & \lambda \times [a,b] & \left\{
\begin{array}{l}
 \, \bigl[ \lambda \times a, \lambda  \times b\bigr] \text{ if } \lambda \geq 0 \\[2pt]
 \, \bigl[ \lambda \times b, \lambda  \times a\bigr] \text{ if not}
\end{array}
\right.
\\
\hline
\rule{0pt}{2.3ex}
\text{general multiplication} & [a,b] \times [c,d] & \bigl[\min(ac,ad,bc,bd),\max(ac,ad,bc,bd)\bigr]\\[2pt]
\hline
\text{inverse} & 1 \,/ \,[c,d] & \left\{ 
\begin{array}{ll} 
\top & \text{ if } c \leq 0 \leq d \\ 
\, \bigl[1/d,1/c\bigr] & \text{ if } 0 < c \leq d \\[2pt]
\, \bigl[1/c,1/d\bigr] & \text{ if } c \leq d  < 0 \\
\end{array}
\right.\\

\hline
\end{array}
\]
The subtraction (resp. division) operators are obtained by adding the opposite (resp. by multiplying by the inverse) of the interval given as a second operand. We can combine the three first operations (addition, opposite and scalar multiplication) to compute the multiplication of an abstraction $X \in \mathcal{P}(\RR^d)$ by a weight matrix $W \in \RR^{p\times d}$ by decomposing $W = W^+ + W^-$ between positive and negative weights and then computing $W X  = W^+ X + W^- X$. In \pyrat, this is used to optimise the computation done on the Box domain for large matrix multiplication, often found in neural networks.

Moreover, note that the abstraction of any increasing function (resp. decreasing function) $f$  is simply $f([a,b]) = [f(a),f(b)]$ (resp. $f([a,b]) = [f(b),f(a)]$). For example, the ReLU function defined as $f(x)= \max(0,x)$ is increasing, therefore its abstraction for an interval [a,b] is:
\[
\text{ReLU}([a, b]) = \max(0,[a,b]) = [\max(0,a), \max(0,b)]
\]
Nevertheless, it is also possible to abstract periodic functions or non monotonic functions such as $\cos$, $\sin$ or sigmoid using interval arithmetic.

Overall, the Box domain is very efficient as it allows to compute reachable bounds for any operation with the same complexity as that operation (on average only 2 operations are needed, one for the lower bound and one for the upper bound of the interval). However, this domain often lacks precision, \eg if $x \in [-1, 1]$ then a simple operation like $x - x = [-1, 1] - [-1, 1] = [-2, 2] \neq 0$ will produce a wider result than expected. Due to the nature of the Box domain, they do not capture the relations between the different variables or inputs of a problem and so may fail to simplify such equations. In the context of neural networks, there are numerous relations between the inputs due to the fully connected or convolutional layers. Therefore, more complex abstract domains that capture the relations between variables such as zonotopes will be presented.

\subsection{Zonotopes}

A \emph{zonotope} $x$ in $\RR^d$, with $d \in \mathbb{N}$ the dimension of the zonotope, is a special case of a convex polytope \cite{goubault2009}; with the particularity of having a central symmetry. It is formally defined as the weighted Minkwoski sum over a set of $m$ \emph{noise symbols} $(\epsilon_1, \dots \epsilon_m) \in [-1, 1]^m$, in other words, for every dimension $1 \leq i \leq d$ we have :
\[
x_i = \left\{\mathop{\sum}_{j=1}^{m} \Alpha_{i,j} \epsilon_j + \beta_i \bigg|\: \epsilon \in [-1, 1]^m \right\}  
\]
where $\Alpha_{i,j}$ are real numbers, called generators, and $\beta_i$ is a real number, called center. This sum can be represented in vector form $\Alpha \mathbf{\epsilon} + \mathbf{\beta}$ where $\Alpha$ is a matrix of dimension $d \times m$ and $\mathbf{\epsilon}$ is a vector of dimension $m$; $\mathbf{\beta}$ is also a vector of dimension $d$.
Since the \emph{zonotope} $x$ is solely characterized by the matrix $\Alpha$ and the vector $\beta$, we use $\langle \Alpha,\:\beta\rangle$ to represent it. 

Unlike boxes, there is no general algorithm to abstract any set $ X \in \mathcal{P}(\RR^d)$ directly into a zonotope. However, if anything else fails, there is always the possibility to first abstract $X$ by a box and then create a zonotope from the box. For each dimension $i$, given a box $[l_i, u_i]$, we can create the zonotope as such:

$$x_i = \frac{u_i - l_i}{2} \epsilon_i + \frac{u_i + l_i}{2}$$ 

with $\epsilon \in [-1, 1]^d$, d new noise symbols.

Moreover, it is easy to obtain a box containing all the possible values taken by the zonotope, through a concretisation operation. For each dimension $i$, the bounds of the box will be given by:
\[
\bigl[\underline{x}_i, \overline{x}_i\bigr] = \bigl[\min x_i, \max x_i\bigr] = \bigl[\beta_i - |\Alpha_i|, \beta_i + |\Alpha_i|\bigr] \text{ with } |\Alpha_i|= \mathop{\sum}_{j=1}^{m} |\Alpha_{i,j}|
\]
Concretising the zonotope into a box allows to compare the box obtained to the safe space defined by the property and thus check whether it holds or not. 

For most operations present in a \nn, the arithmetic of zonotope is simple as it relies on affine arithmetic \cite{comba1993ne}. Indeed most of the \nn operations are affine functions for which a zonotope can perfectly capture their results. The table below shows the addition, the opposite and the scalar multiplication on a zonotope.

\[
\begin{array}{|l|c|l|}
\hline
\text{addition} & \mzono{\Alpha}{\beta} + \mzono{\Alpha'}{\beta'} & \mzono{\Alpha + \Alpha'}{\beta + \beta'}\\
\hline
\text{opposite} & -\, \mzono{\Alpha}{\beta} & \mzono{-\Alpha}{-\beta}\\
\hline
\text{scalar multiplication} & \lambda \times \mzono{\Alpha}{\beta} & \mzono{\lambda\times\Alpha}{\lambda\times\beta}\\
\hline
\end{array}
\]

\begin{figure}[ht]
    \centering
    \includegraphics[width=2in]{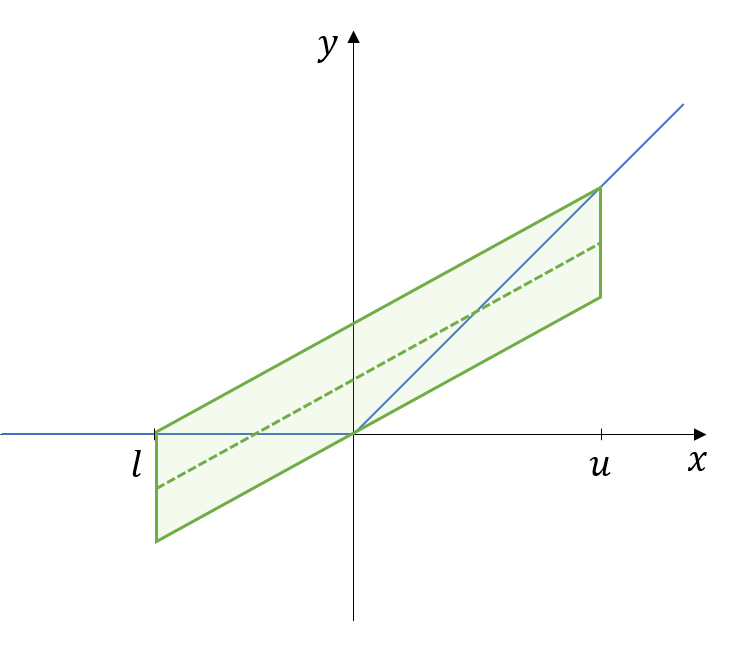}
    \includegraphics[width=2in]{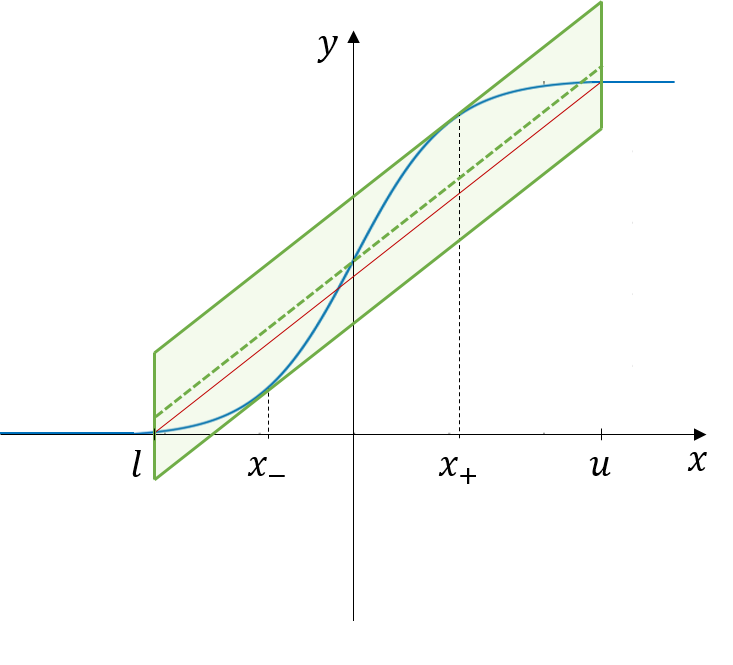}
    \includegraphics[width=2in]{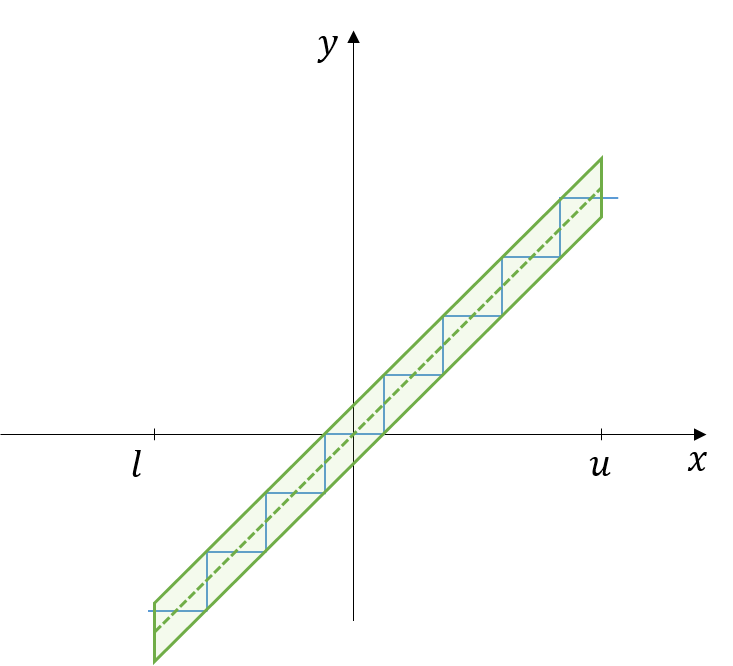}
	\caption{Activation functions and their \zono abstraction (Relu, Sigmoid, Cast)}
	\label{fig:abstractions}
\end{figure}

For non-linear operations, such as the activation functions or multiplication or division, we need to use an abstraction of the operation to handle them as the affine arithmetic cannot represent these operations otherwise. Thus we are constructing an over-approximation of the reachable output of the layer.
We present quickly the abstraction for the ReLU, Sigmoid and Cast functions as shown in ~Figure \ref{fig:abstractions}. The abstraction are constructed neuron-wise, \ie independently for each element or neuron of a given layer. In that sense, we can consider a single dimension input zonotope $x$ living in $[l, u] \subset \RR$ for which we want to compute the output zonotope $y$ through a non-linear activation function. In turn, this can be generalised to a multi-dimension zonotopes.

\paragraph{\relu} 
$$ReLU(x) = \max (x, 0)$$
The \relu function is thus a piecewise linear function, if $x < 0$ then $\relu(x) = 0$ otherwise $\relu(x) = x$. As such, it is not a linear function for all values of $l$ and $u$ but is linear if x is not in the interval $[l, u]$. Thus we consider three cases for the \relu abstraction. The first two cases consider a stable or linear \relu, where the interval is either fully positive or fully negative. If $u < 0$, then we have $y = 0$ and if $l > 0$, the zonotope is fully positive and $y = x$.

The last case is the general case if $l < 0 < u$ where the \relu is said to be unstable. As it not linear, we need to provide a linear abstraction of the \relu as in DeepZ \cite{deepzono}. This abstraction $\relu^\sharp$ is parameterised by $a = \frac{u}{u - l}$ and $b = -\frac{ul}{u - l}$ and requires adding a new noise symbol to the zonotope $\epsilon_{m+1}$:
$$y = ReLU^\sharp(x) = ax + \frac{b}{2} + \frac{b}{2}\epsilon_{m+1} = \left\langle\left(\begin{array}{c}
     a \times \Alpha\\
     \frac{b}{2} 
\end{array}\right), a\times\beta + \frac{b}{2}\right\rangle$$

\paragraph{Sigmoid}
$$\text{Sigmoid}(x) = \frac{1}{1 + e^{-x}}$$

For the Sigmoid function, we build on the work presented in \cite{signed2013} to construct our abstraction. First, we compute $x_-$, (respectively $x_+$) the point where the Sigmoid is the furthest below (respectively above) the slope between Sigmoid$(l)$ and Sigmoid$(u)$. If $l > 0$, we define $x_- := l$ and, similarly, if $u < 0$, $x_+ := u$.
With these points, we can obtain the distance from the slope to the Sigmoid and thus create our \zono abstraction:
$$y = \text{sigmoid}^\sharp(x) = ax + \frac{b}{2} + \frac{\delta}{2}\epsilon_{m+1} = \left\langle\left(\begin{array}{c}
     a \times \Alpha\\
     \frac{b}{2} 
\end{array}\right), a\times\beta + \frac{b}{2}\right\rangle$$
with 
\[
\begin{array}{cl}
    a &= \dfrac{\text{Sigmoid}(u) -\text{Sigmoid}(l)}{u - l}\\
    b &= -al + \text{Sigmoid}(l) + \dfrac{\text{Sigmoid}(x_+) + \text{Sigmoid}(x_-) - a(x_+ + x_-)}{2} \\[7pt]
    \delta &= \dfrac{\text{Sigmoid}(x_+) - \text{Sigmoid}(x_-) - a(x_+ - x_-)}{2}
\end{array}
\]

Here a new noise symbol $ \epsilon_{m+1}$ will always be introduced for all values of $l$ and $u$ as the function is not piece-wise linear.

\paragraph{Cast}

The Cast layer and its variants, such as Ceil or Floor, are operations used to change the data type from float32 to int8 for example or to replace the activation function in some networks. These operations are not linear and thus need an abstraction. We only detail the abstraction of the Cast operation here but Floor and Ceil are similar. 
Here, we have two cases for this abstraction due to its piece-wise linear nature: if $l$ and $u$ are in the same linear part, then $y = \text{cast}(l)$; otherwise we add a new noise symbol $\epsilon_{m+1}$ with a linear abstraction of the floor function.
$$y = cast^\sharp(x) = x + 0.5\epsilon_{m+1} = \left\langle\left(\begin{array}{c}
     \Alpha\\
     0.5 
\end{array}\right), \beta + 0.5\right\rangle$$

\paragraph{Complexity reduction for Zonotopes}
As the number of abstraction tend to grow in large networks, the number of noise symbols introduced in the network can also grow significantly in the analysis. These symbols can in turn slow down the analysis a great deal as the complexity of the operations on zonotopes is at least $\mathcal{O}(m d)$ with $d$ the number of dimension of the zonotope and $m$ the number of noise symbol in the zonotope. Multiple heuristics to reduce the number of noise symbols have been implemented in \pyrat; for example merging some noise symbols in order not to exceed a set number of noise symbols. Criteria such as the one presented in \citet{kashiwagi2012algorithm} can be used to choose the noise symbols to merge. Moreover as we merge these noise symbols, we create an over-approximation of their values so that the analysis with \pyrat remains correct although it does loose some precision. Different parameters allow the user to tune this reduction efficiently depending on the network used.

\subsection{Constrained Zonotopes}
Building on the work in \citet{scott_constrained_2016}, we extend our \zono domain by adding linear constraints shared over the $d$ dimensions of the \zono. These constraints allow for more precise abstractions of non-linear functions, reducing the over-approximation introduced during an analysis. At the same time, these constraints can be used to split functions into multiple parts for greater precision in a \bab approach as this will be detailed in section \ref{sec:bb}. Instead of the equality constraints used in \citet{scott_constrained_2016}, we use inequality constraints as the equality constraints are not meaningful in the context of our abstractions. 

More formally a \conz $x = (x_1, ..., x_d) \in \RR^d$ with $K$ constraints and $m$ \emph{noise symbols} is defined by:
\[
x_i =  \left\{\mathop{\sum}_{j=1}^{m} \Alpha_{i, j} \epsilon_j + \beta_{i}  \,\Bigg|\, \begin{array}{c}
     \epsilon \in [-1, 1]^m  \\
     \forall k \in \{1, ..., K\}, 
    \mathop{\sum}_{j=1}^{m} A_{k, j} \epsilon_j + b_k \geq 0
\end{array}
\right\}
\]
where $A_{k, j} \in \RR$ and $b_k \in \RR$. Note that $A$ and $b$ are not indexed by $i$, implying that the constraint are shared for all $i$. At the start of the analysis, the \conz is  unconstrained with K = 0 and we can add constraints at different layers.

\begin{figure}[h!]
    \centering
    \includegraphics[width=0.5\linewidth]{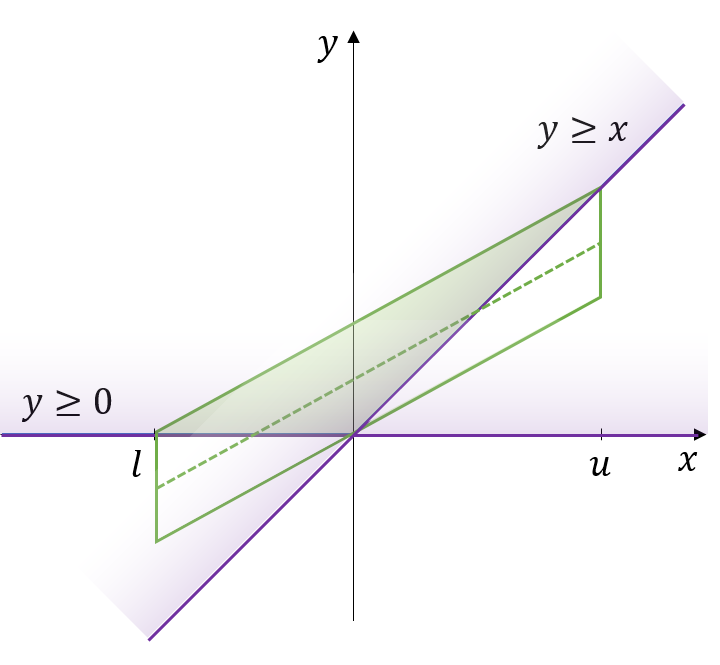}
    \caption{Abstraction of the \relu with the \conz domain}
    \label{fig:conz_relu}
\end{figure}

An example of this is shown in Figure~\ref{fig:conz_relu} with the ReLU abstraction when using the \conz domain. The linear abstraction remains unchanged (in green) but two constraints (in purple) are added $y \geq x$ and $y \geq 0$ reducing the approximation introduced. These constraints are added neuron-wise for each unstable neuron, as such for a \relu layer of dimension $n$, we add $2p$ constraints with $0 \leq p \leq n$ the number of unstable \relu neurons.

While the \zono could easily be concretised into a Box through a simple linear operation, for a tight concretisation from \conz to Box, we have to make use of the linear constraints of the \conz. For this, we need to solve a system of linear equations with linear inequality constraints. The lower bound for all $i \in \{1, ..., d\}$ is given by:
$$
\min_{\substack{\epsilon \in [-1, 1]^m \\ \left(\sum_{j=1}^m A_{k,j} \epsilon_j + b_k \geq 0\right)_{k=1, ..., K}}} \, \sum_{j=1}^m \Alpha_{i, j} \epsilon_j + \beta_{i}
$$
and the upper bound by:
$$
\max_{\substack{\epsilon \in [-1, 1]^m \\ \left(\sum_{j=1}^m A_{k,j} \epsilon_j + b_k \geq 0\right)_{k=1, ..., K}}} \sum_{j=1}^m \Alpha_{i, j} \epsilon_j + \beta_{i}
$$
To solve these optimisation problems, a classical solution is to use a Constraint Programming solver. However, as the size of the neural networks grows and more constraints are introduced, these solvers have difficulties scaling up. On top of this, solvers are often unsound w.r.t. real arithmetic and do not match with the verification approach in \pyrat to have a sound approximation of the neural network's output. With this in mind, we chose to build a sound method to solve this. Thus, we first rewrite the equation into its Lagrangian counterpart before solving for the dual which will always yields sound bounds. Then we can make use of the techniques developed in optimization theory to solve our problem. 
More formally, to obtain the lower bound, we need to solve for $i \in \{1, ..., d\}$
$$
\min_{\substack{\epsilon \in [-1, 1]^m \\ \left(\sum_{j=1}^m A_{k,j} \epsilon_j + b_k \geq 0\right)_{k=1, ..., K}}} \sum_{j=1}^m \Alpha_{i, j} \epsilon_j + \beta_{i}
$$
which is equivalent to solving the following min-max problem by introducing a Lagrange multiplier $\lambda_k$ with $ 0 \leq k \leq K$ for each constraint of the \conz:
$$
\min_{\epsilon \in [-1, 1]^m} \max_{\lambda \in \RR_+^K} \sum_{j=1}^m \Alpha_j \epsilon_j + \beta - \sum_{k=1}^K \lambda_k \left(\sum_{j=1}^m A_{k,j} \epsilon_j + b_k \right)
$$
where we dropped the subscript $i$ for readability. Since we have strong duality because of the linearity of the objective and the compactness of the optimization variables, we can reverse the order of the min and the max and obtain the same bound. Thus, yielding:
\begin{align*}
\max_{\lambda \in \RR_+^K} \min_{\epsilon \in [-1, 1]^m} \sum_{j=1}^m \Alpha_j \epsilon_j + \beta - \sum_{k=1}^K \lambda_k \left(\sum_{j=1}^m A_{k,j} \epsilon_j + b_k \right) &= 
\max_{\lambda \in \RR_+^K} - \sum_{j = 1}^m \left| \Alpha_j - \sum_{k=1}^K \lambda_k A_{k,j} \right| - \sum_{k=1}^K \lambda_k b_k + \beta
\end{align*}
Hence, we have
$$
\min_{\substack{\epsilon \in [-1, 1]^m \\ \left(\sum_{j=1}^m A_{k,j} \epsilon_j + b_k \geq 0\right)_{k=1, ..., K}}} \sum_{j=1}^m \Alpha_{j} \epsilon_j + \beta = \max_{\lambda \in \RR_+^K} - \sum_{j = 1}^m \left| \Alpha_j - \sum_{k=1}^K \lambda_k A_{k,j} \right| - \sum_{k=1}^K \lambda_k b_k + \beta
$$
One can observe that taking any $\lambda \in \RR^K_+$ will yield a valid lower bound of the original problem. Thus, we can use classical gradient ascent algorithm like SGD~\cite{SGD} or Adam~\cite{Adam} to find the optimum. This approach allows us to have a good quality bound in an efficient manner. 

\subsection{Hybrid Zonotopes}
Introduced in the field of closed loop dynamical systems by \citet{hz} and then extended to neural networks by \citet{hz-2zono} and \citet{hz-3zono}, an \hybz can be seen as the union of \conzs. Over $d$ dimensions, we define an \hybz $x_i$ for all dimension $i \in \{1, ..., d\}$ with $K$ constraints similarly to other \zonos with $m_c$ \emph{continuous noise symbols} and $m_b$ \emph{binary noise symbols}:
\[
x_i = \left\{\mathop{\sum}_{j=1}^{m_c} \alpha_{i, j} \epsilon^c_j +\mathop{\sum}_{j=1}^{m_b}  \gamma_{i, j}\epsilon^b_j + \beta_{i} \,\Bigg|\, 
\begin{array}{c}
     \epsilon^c \in [-1, 1]^{m_c} \\
     \epsilon^b\in \{-1, 1 \}^{m_b} \\
     \forall k \in \{1, ..., K\}, 
    \mathop{\sum}_{j=1}^{m_c} a_{k, j} \epsilon^c_j + \mathop{\sum}_{j=1}^{m_b }b_{k, j} \epsilon^b_j + c_k = 0
\end{array}
    \right\}
\]
where $\gamma_{i, j} \in \RR$ is called binary generators and $a_{k, j}$, $b_{k, j}$, $c_k \in \RR$.

The union of \conzs is here defined through the binary noise symbols that can only take the discrete values of $-1$ or $1$. The \hybz is thus the union of $2^{m_b}$ \conzs. The constraints used here are equality constraints as they allow to define the intersection between two \conz (defined by \cite{scott_constrained_2016}) which is needed for the \hybzs.

The arithmetic is the same as classical \zono. Nevertheless, an \hybz has the advantage of being able to exactly represent any piece-wise linear function such as the ReLU or the cast function. With them, we can thus obtain an exact representation of any neural network for which the activation function are piece-wise linear (which comprise a good part of the currently used networks). 

\begin{figure}[h]
    \centering
    \includegraphics[width=2.5in]{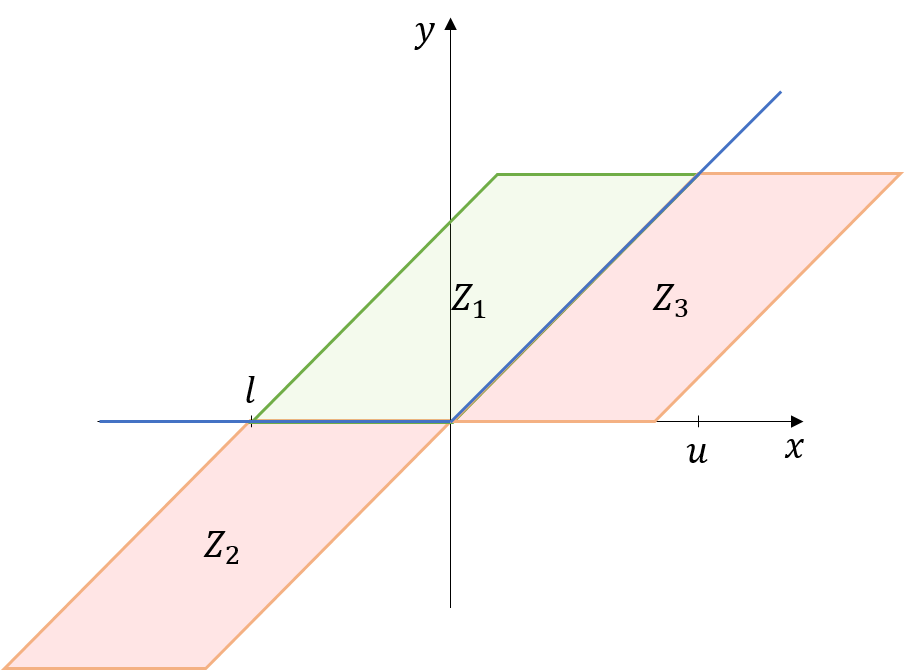}
    \includegraphics[width=2.5in]{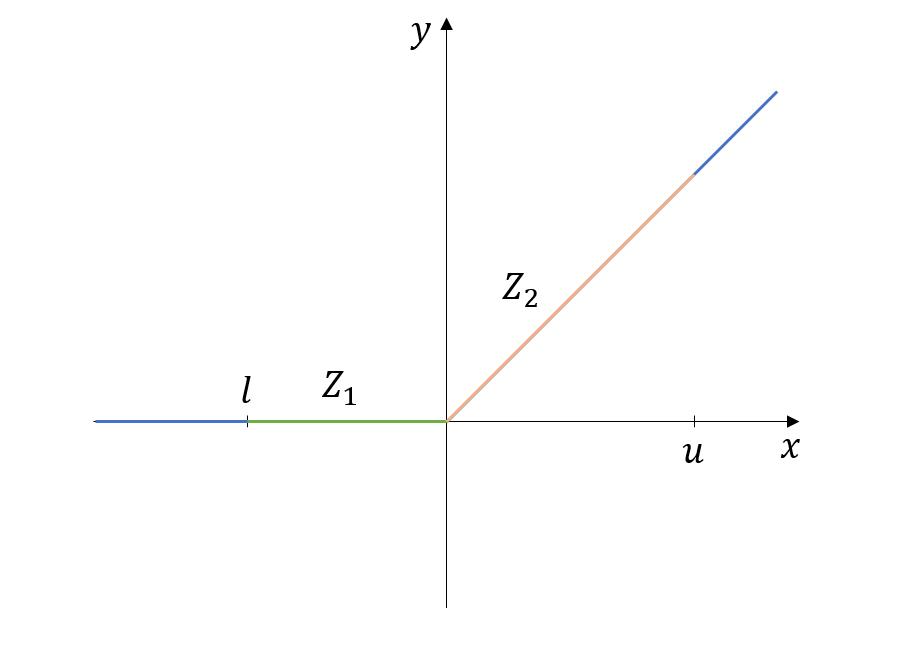}
	\caption{\textit{Hybrid zonotope} abstraction of the \relu function. $H = Z_1 \cap (Z_2 \cup Z_3)$ \cite{hz-3zono} on the left. $H = Z_1 \cup Z_2$ from \citet{hz-2zono} on the right.}
	\label{fig:abstractions_hz}
\end{figure}

We will now briefly introduce how to exactly represent the ReLU function with an \hybz. For this, two representations of the ReLU function are proposed by \citet{hz-3zono} and \citet{hz-2zono}. Figure~\ref{fig:abstractions_hz} shows both representation. We refer the reader to the respective paper for the exact equation of their representations. With these representations, we obtain an \hybz representing the input-output set of the ReLU function for each dimension $0 \leq i \leq d$:

$$H_i = \{(x_i, y_i) \in \RR^2\ |\ y_i = ReLU(x_i), x_i \in [l_i, u_i]\}$$

The cartesian product over all dimensions returns the exact input-output set of the ReLU for all dimension as an \hybz $H = \bigtimes_{i=1}^d H_i$. Then, the intersection of this \hybz $H$ with $X$ the input of the ReLU also represented by an \hybz will give us the exact output of the ReLU function $Y = \text{ReLU}^\sharp(X) = H \cap_{[\mathbb{I}_d 0]} X$ with the intersection defined in~\citet{hz}.

Considering the input dimension at the ReLU layer is $d$, this abstraction will add $4d$ new continuous noise symbols, $d$ binary noise symbols and $3d$ constraints. At this step, we avoid an exponential increase of the constraints and noise symbols by remaining linear in our representation.

Repeating this for all the \relu in the network will give an exact representation of the network. This can allow one to reuse this representation over multiple inputs instead of doing an analysis for each input.
Nevertheless, we need to be able to find the bounds of the output, \ie concretise it into a box, or prove that its intersection with the negation of the safe space is empty in order to assess whether the property holds. Considering the binary noise symbols in the constraints, this means solving a mixed integer linear programming (MILP) problem. Overall, the \hybz approach will provide a MILP encoding of the network and thus transfer the complexity to a MILP solver.
Given the difficulty of solving this MILP problem, the \hybz abstraction only scale on small networks for now but various solutions are considered for future improvements.

\subsection{Other domains}

\pyrat also implements a variety of other domains such as polytopes from \citet{deeppoly}, polynomial zonotopes \cite{Kochdumper_2023} or small sets. \pyrat also provides interfaces to easily extend its support to new domains either for quick proof of concept or full implementation.

\section{PyRAT}
In this section, we will present the main functionalities of \pyrat as a tool to verify neural networks. More specifically, we will detail how \pyrat applies abstract domains to perform a reachability analysis and proves safety properties.

\begin{figure}[h!]
    \centering
    \includegraphics[width=1\linewidth]{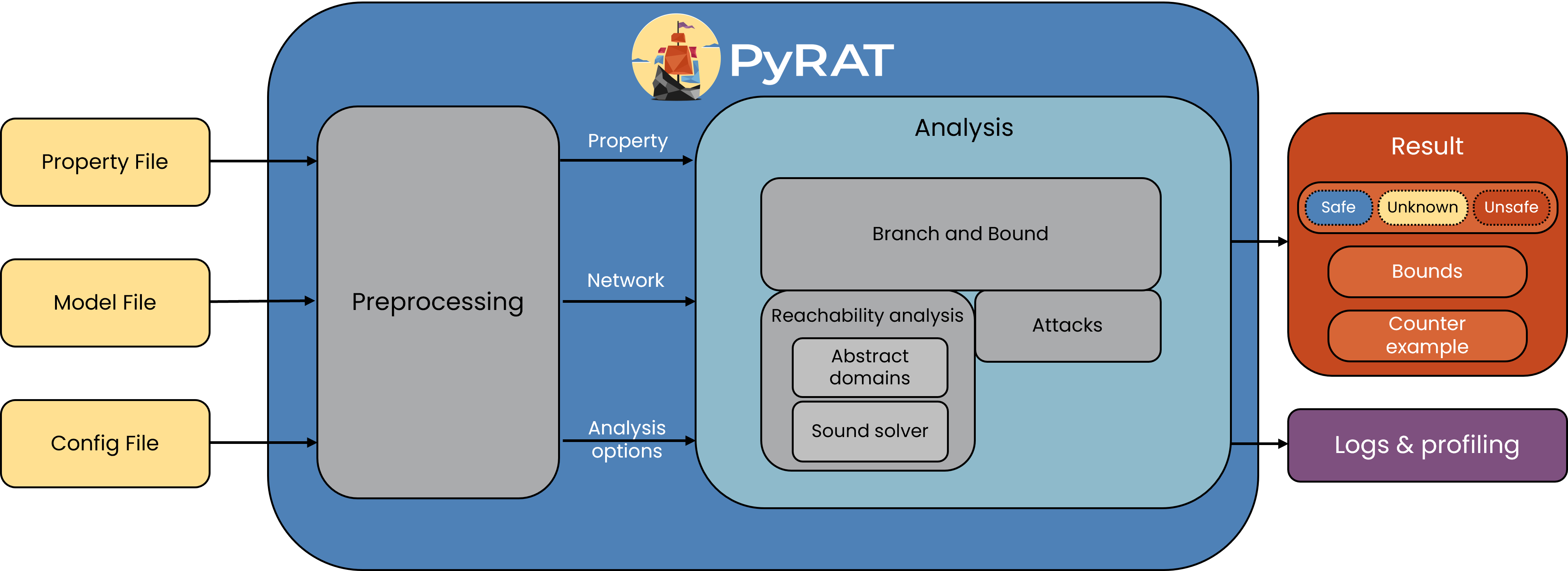}
    \caption{High-level architecture of \pyrat}
    \label{fig:highlevel}
\end{figure}

\subsection{Network and property parsing}

The first step performed in \pyrat before any analysis is the parsing of the inputs, \ie the \nn to analyse and the property to verify. For the \nn, \pyrat supports the ONNX standard as its main input but it can also parse, though with a more limited support, Keras/Tensorflow or PyTorch models. 
These networks are transformed in \pyrat's own internal representation which will then serve to apply all the operations in the network on our abstract domains.

\begin{figure}[h]
    \centering
    \includegraphics[height=0.7in]{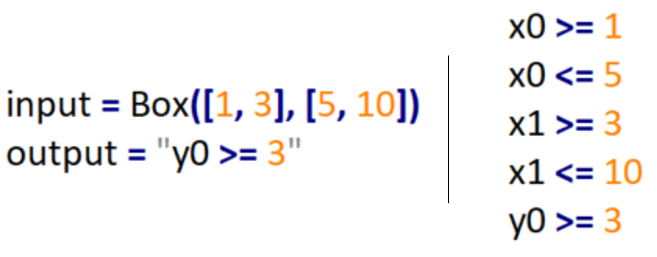}
	\caption{Equivalent properties for \pyrat in Python (left) and in a text file (right).}
	\label{fig:property}
\end{figure}

For the parsing of the property, \pyrat supports the \vnnlib\footnote{\url{https://www.vnnlib.org/}} property specification language which is itself based on the SMT-LIB format. At the same time, \pyrat supports its own specification format in textual mode or through a Python API as shown in Figure~\ref{fig:property}.

The simplest way to define properties in \pyrat is through scalar constraints on each input, i.e. an interval for each input, and a condition to be respected on the outputs which can be conjunctions, disjunctions or negations. More complex properties can be defined using both inputs and outputs at the same time but they remain limited in expressivity. Outputs and inputs are numbered according to their order in a flattened array, if the inputs are multidimensional.

Once both the property and the network are parsed, \pyrat can perform some simplifications on the network to facilitate the verification. These may include removing redundant transpose layers, removing unneeded softmax, fusing matmul and add, etc. These operations are made for the sake of the analysis and remain conservative in the sense that they do not alter the result of the network inference w.r.t. the property. Two simplifications are further presented here:

\paragraph{Adding a layer}

When using relational domains such as the Zonotope domain, \pyrat is able to add an additional layer at the end of the network to make full use of the relations for certain properties. For example, consider the property "$y_0 \leq y_1$", and the output to be over-approximated by:
\begin{align*}
y_0^\sharp &= 3 \epsilon_0 + 2 \epsilon_1 - 2 \\
y_1^\sharp &= \epsilon_0 + 2 \epsilon_1\,.
\end{align*}
Evaluated separately with intervals, we have $y_0 \in [-7, 3]$ and $y_1 \in [-3, 3]$, thus we cannot conclude on the property. However, $y_0 \leq y_1 \Leftrightarrow y_0 - y_1 \leq 0$ and if we compute the difference between the \zonos we have $z_0 = y_0^\sharp - y_1^\sharp = 2 \epsilon_0 - 2 \in [-4, 0]$ which verifies the property.

As such, \pyrat will add a matrix multiplication at the end of the network to create the new variables $z_i$ and rewrite the property in terms of these $z_i$.

\paragraph{Maxpool to \relu}

As introduced in DNNV \cite{Shriver_2021}, \pyrat is able to simplify the neural network for verification by transforming maxpool layers which are harder to verify into convolutions and \relu layers. On top of the initial simplification by DNNV, we propose an improvement to the simplification. Indeed, the initial simplification relies on the fact that: $$\max{(a, b)} = \relu(a-b) + \relu(b) - \relu(-b)$$ 
Thus any maxpool layer can be transformed in a succession of Convolution (to extract the $a-b$, $b$ and $-b$) and of \relu. For a maxpool of kernel $k$ this results in approximately $2log_2(k)$ convolutions followed by \relu layers. 
While it allows to verify more precisely a maxpool layer this simplification is not without issues: we must compute additional convolutions, and do the abstraction of three new non-linear \relu functions each time. For the latter, when using the \zono domain this will introduce new noise symbols if $\gamma(b) = [l, u]$ contains $0$ twice, once for $\relu(b)$ and once for $\relu(-b)$.
An improvement on this simplification is instead to rely on $$\max{(a, b)} = \relu(a - b) + b$$ and use a convolution to obtain $a - b$ and $b$ followed by a \relu only on $a-b$ and an Add layer to obtain the output. This will reduce the convolution size by a third and avoid unnecessary over-approximation. 

\begin{figure}[h]
    \centering
    \includegraphics[height=0.7in]{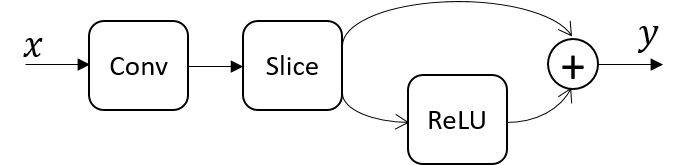}
	\caption{Maxpool layer transformed in \pyrat.}
	\label{fig:maxpool}
\end{figure}

\subsection{Reachability analysis in \pyrat}
\label{sec:indiv_an}
After these steps of parsing and preprocessing, \pyrat will create the abstract domains from the input box specified in the property. It will then apply all the operations of the network on the input box and the abstract domains following the graph execution order. These operations will be handled according to the specific arithmetic of each domain and abstracted as detailed in the previous section for non-linear functions. 

For each analysis, \pyrat is able to use multiple domains at the same time in order to benefit from the strengths of each of them and complement their approaches. By default, the Box domain which has a relatively low cost will always be used. It is useful for layers such as the \relu layer where the Box domain bounds are exact and thus, we can compute the intersection of the Box domain with other domains that require over-approximating the output of the \relu to improve the precision of our result.

In order to provide a fast implementation on the highly dimensional neural networks, \pyrat implements its abstract domains  using NumPy arrays or PyTorch tensors. A choice between NumPy or PyTorch can be done automatically in function of the input dimension. Experimentally, NumPy is faster on smaller size matrices than PyTorch. On large networks, one can also leverage GPU computation through PyTorch with \pyrat, thus scaling to much larger networks. Nevertheless, using GPU will lose the soundness of \pyrat w.r.t. real arithmetic as it will not be able to change the rounding mode of floating point variables. 
Additional computation libraries can also be implemented in \pyrat through predefined APIs.

While the primary objective of \pyrat remains to prove that the property holds, it can happen that the property does not hold. Such cases tend to be hard to handle with abstract interpretation alone as it over-approximates the outputs. To handle this, \pyrat will look for counter example on top of the reachability analysis performed. Random points and adversarial attacks will be generated before the analysis to try to falsify the property by launching inferences with the original model. If a counter-example is found, \pyrat will return it and conclude the analysis as the property is falsified. 
On the other side, if our reachability analysis manages to prove that a certain input Box is false, a counter example will be selected in the input Box to return to the user. 
This counter-example search is tunable with the adversarial attack parameters and can be deactivated if needed.

At the end of its forward pass on the network, \pyrat will obtain output abstract domains over-approximating the true possible outputs of the network and will evaluate the property on these outputs. As \pyrat is always over-approximating the possible solutions, it is \textbf{correct} and will not say a property is verified or falsified if this is not true. However, the abstractions used may be too rough and \pyrat will not be able to conclude on them leading to an unknown output. As such the possible outputs for \pyrat are:

\begin{itemize}
\item "True", the property is verified.
\item "False", the property is false and a counter-example is provided.
\item "Unknown", \pyrat could not verify nor falsify the property with the options and abstractions used.
\item "Timeout", the property could not be verified nor falsified under the given time.
\end{itemize}

When using branch and bound methods on \relu neural networks, as described by Section~\ref{sec:bb}, \pyrat will be \textbf{complete} in the sense that provided an infinite timeout it will always give an answer either "True" or "False".

\subsection{\Bab}
\label{sec:bb}
As described in Section~\ref{sec:indiv_an}, individual analysis in \pyrat can result in "Unknown" evaluation. This happens because the abstract domains are too loose to represent precisely the output, for this reason, we use \bab techniques which rely on multiple analyses to increase the precision. \\
This method comes from the field of mathematical optimization \cite{bab_origin} which consists of breaking the original optimization problem into smaller optimization problems and offers the possibility to prune sub-problems if they cannot improve the optimal bound. 
In the context of \nn verification, if the analysis cannot prove or falsify the property with the chosen abstract domain, \pyrat offers the possibility to perform multiple analyses to give more accurate results by partitioning the verification problem. There are multiple ways to perform this partitioning and all can be described in the \bab framework~\cite{bunel2020branch}. \pyrat implements two well known \bab techniques: \bab on the inputs or on the \relu nodes.

\paragraph{\Bab on input}
An intuitive way to reduce the over-approximation on the output is to partition the input problem recursively until every sub space is proven or until one of the sub space invalidates the property (see for example ~\cite{reluval, marabou}). This is best described in algorithm \ref{alg:BaB_input}. 

\begin{algorithm}
	\caption{\bab on input in \pyrat}\label{alg:BaB_input}
	\begin{algorithmic}[1]
		\Require neural network $N$, output property $p$, input lower bound~$\underline{x}$, input upper bound $\overline{x}$, partitioning number $k$.
		\vspace{2.5pt}
		\State $v \,\gets$ [($\underline{x}, \, \overline{x}$)]
		\label{line:v}
		\Comment{\parbox[t]{.5\linewidth}{$v$ contains the bounds of each unknown problem}}
		\While{$v$ is not empty}
		\State $\underline{x}, \, \overline{x} \gets $
		$v$.pop()
		\State $r,\,\alpha\,\gets$ \Call{Analyse}{$N,\, p,\, \underline{x}, \overline{x}$}
		\label{line:analyse}
		\If{$r$ is False} return False
		\label{line:cex}
		\EndIf
		\If{$r$ is unknown}
		\State $j \gets $ \Call{ChooseDim}{$\alpha$}
		\label{line:choose}
		\State $splits \gets $ \Call{Split}{$\underline{x}, \, \overline{x}$, $j$, $k$}
		\label{line:split}
		\Comment{\parbox[t]{.5\linewidth}{Split into $k$ parts the input bound at dimension $j$}}
		\State Append $splits$ to $v$
		\label{line:to_split}
		\EndIf
		\EndWhile
		\State\Return True \label{line:true}
	\end{algorithmic}
\end{algorithm}

First, the algorithm creates a list that will store each input problem that has to be verified (line \ref*{line:v}). One of the problems is selected and analysed by \pyrat (line \ref*{line:analyse}). If the analysis result is unknown, it chooses the input dimension that is deemed the most important to split upon. \pyrat estimates this importance according to the influence, $\alpha$, of the input dimension on the property (line \ref*{line:choose}). Then, the input bounds are partitioned in $k$ equal parts along the chosen dimension $j$ (line \ref*{line:split}).
These new input bounds define $k$ sub-problems that are then added to the list of sub-problem to verify. \\
This algorithm ends either when all of the sub-problem are verified (line \ref*{line:true}) or when a counter-example is found for a sub-problem (line \ref*{line:cex}). Since each sub-problem is a subset of the original problem, the algorithm yields True when the property is verified and False when it is not.

The core of the algorithm lies in the choice of the dimension to split, and therefore, the quality of the estimation $\alpha$. To estimate the importance of the input \wrt property, \pyrat essentially relies on the relation of the available abstract domain between the input and the output $\alpha_{i,j}$ and the width of the input $x_i \in [\underline{x_i}, \overline{x_i}]$:
$$
\alpha_i \,:=\,\sum_{j \in \text{output dim}} \alpha_{i,j} (\overline{x_i} - \underline{x_i})
$$
For a more in-depth explanation of the heuristic, refer to~\citet{reciph}. 

This \bab approach on the input performs well on low dimensional input problem but has a low efficiency when the input space get bigger, \eg for images. Indeed, with more inputs, the number of division to perform grows exponentially to achieve an increase in precision as the input are often very interconnected and multiple inputs need to be split. To resolve this issue, \bab on \relu function has been developed. 

\paragraph{\Bab on \relu}
The idea of branching on \relu comes from \tool{Reluplex}~\cite{reluplex} and is now implemented in every state-of-the-art verifier such as \tool{alpha-beta-CROWN}~\cite{betacrown} and \tool{MN-BaB}~\cite{mn-bab}. Branching on a \relu means we enforce the input of a specific \relu to be either positive or negative which transform the \relu operation in a simple linear operation: the $0$-function if the input is negative and the identity function if the input is positive. Providing enough time for the analysis, this approach is complete since our abstract domains represent exactly linear functions. In addition, branching on \relu nodes allows to split multiple input dimensions at once since \relu nodes are often dependant on multiple inputs. Contrary to branching on inputs this allows to address higher dimension problems. 

Conceptually, the algorithm for splitting on \relu is quite similar to the one for input splitting: it chooses an unstable \relu in the neural network, it forces the aforementioned \relu to be negative and then analyses the neural network, it then analyses the network with the chosen \relu being positive. Since \relu$(x)$ is the $0$ function for $x \in ]-\infty, 0]$ and the identity for $x \in [0, +\infty[$, partitioning at $x=0$ allows us to represent it exactly with 2 linear functions. The main differences between splitting on input and splitting on \relu are highlighted in Algorithm \ref{alg:BaB_relu}.

\begin{algorithm}
	\caption{\bab on \relu in \pyrat}\label{alg:BaB_relu}
	\begin{algorithmic}[1]
		\Require the neural network $N$, the output property $p$, the input lower bound~$\underline{x}$, the input upper bound $\overline{x}$.
		\vspace{2.5pt}
		\State \diff{$v \,\gets [\;]$}
		\Comment{\parbox[t]{.4\linewidth}{$v$ contains the indices of the \relu to split along with their sign}}
		\While{$v$ is not empty}
		\State $split, \gets \,v$.pop()
		\State \diff{$r, \, \gamma \gets$ \Call{AnalyseSplit}{$N,\, p, \underline{x}, \, \overline{x}, \,split$}}
		\label{bab_rel:analyse}
		\If{$r$ is False} return False
		\label{bab_rel:cex}
		\EndIf
		\If{$r$ is unknown}
		\State \diff{$l,\,j \gets $ \Call{Choose\relu}{$\gamma,\, \neg \,split$}}
		\Comment{\parbox[t]{.4\linewidth}{Chooses \relu $j$ at layer $l$ to split on}}
		\label{bab_rel:choose}
		\State \diff{$split\_n \gets$ \Call{CopyAndAdd}{$split,\, (l, j) \leq 0$}}
		\State \diff{$split\_p \gets$ \Call{CopyAndAdd}{$split,\, (l, j) > 0$}}
		\State Append $split\_n$ and $split\_p$ to $v$
		\label{bab_rel:to_split}
		\EndIf
		
		\EndWhile
		\State\Return True \label{bab_rel:true}
	\end{algorithmic}
\end{algorithm}

The main difference is the replacement of the function \tool{Analyse} by the function \tool{AnalyseSplit}. This function takes into account the splits constraints on the \relu and then applies these constraints to the different abstract domains. 

For example with the \conz domain, if the algorithm has to split \relu $i$ at layer $l$, denote by $x_i^l$ the pre-\relu \conz abstraction and by $y_i^l$ the \relu abstraction of $x_i^l$. 
$y_i^l$ with negative split is represented in the \conz domain as $0$ with the constraint $y_i^l < 0$ added to the existing constraints of the \conz. $y_i^l$ with positive split is represented by $x_i^l$ with the constraint $y_i^l \geq 0$ added.

Similarly to \bab on inputs, the choice of the \relu node to split upon is crucial. The more unstable \relu node there is, the greater the importance of this choice to avoid an exponential increase in the number of sub-problems. Here we want to estimate the influence of the node $i$ of layer $l$ on the outputs. \\

\pyrat relies on the relation $\alpha_{\relu_i, l}$ between the output layer and the $\relu_i^l$ nodes to estimate this influence. For the \zono domains, this stems from the noise symbols added at each unstable \relu node, while we can compute a gradient to the \relu nodes for polyhedra domains. This coefficient is then multiplied by the impact $\delta_+$ on the \relu when fixing it to be positive and $\delta_-$ when the \relu is fixed negative.
$$
\gamma_i^l := (\delta_+ + \delta_-) \times \alpha_{\relu i,l}
$$

This estimate does not take into account the influence of fixing a \relu node on later \relu, \eg it can also fix a \relu node to be positive in the next layer, nor does computes the influence of the newly added constraint as it can also reduce the imprecision on another \relu node of the same layer or of another layer. Nevertheless, this allows a fast estimation of the influence of a \relu node with a good efficiency. Improvements on this estimate will be subject of future works.

\section{Usage}
\pyrat is packaged as a Python 3 module and, once downloaded, can be installed with \texttt{pip} on any system with a Python version superior to 3.10. And while \pyrat is closed-source, it can be made freely available under an academic licence for research and teaching purposes. 

\begin{figure}[h!]
    \centering
    \includegraphics[width=0.7\linewidth]{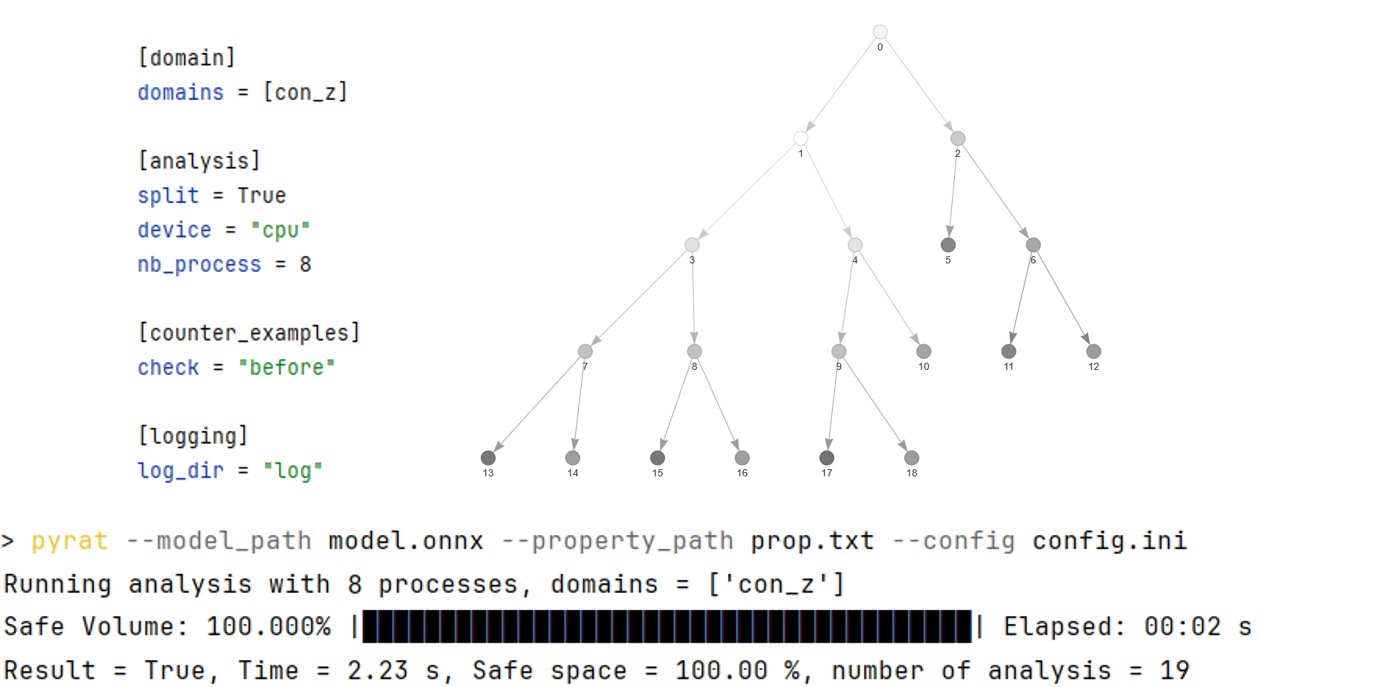}
    \caption{Example of configuration file (top left), command line execution (bottom) and output split tree (top right)}
    \label{fig:config}
\end{figure}
As a Python module, \pyrat can be used directly in Python or from command line. Using \texttt{ConfigArgParse}, it can take arguments from the command line using specific keywords or with a configuration file containing the different parameters.
During the analysis, \pyrat displays information such as the current time taken by the analysis along with the percentage of completion for \bab. Without \bab the output bounds of the analysis will be shown at the end. The Python usage allows one to investigate the results more in details, fetching the values of the different abstract domains used for different layers or the decisions taken during a \bab analysis. Some information can also be logged in files and the \bab decisions can be saved and visualised as a split tree using \texttt{networkx} (see Figure~\ref{fig:config}). 

\begin{figure}[h!]
    \centering
    \includegraphics[width=\linewidth]{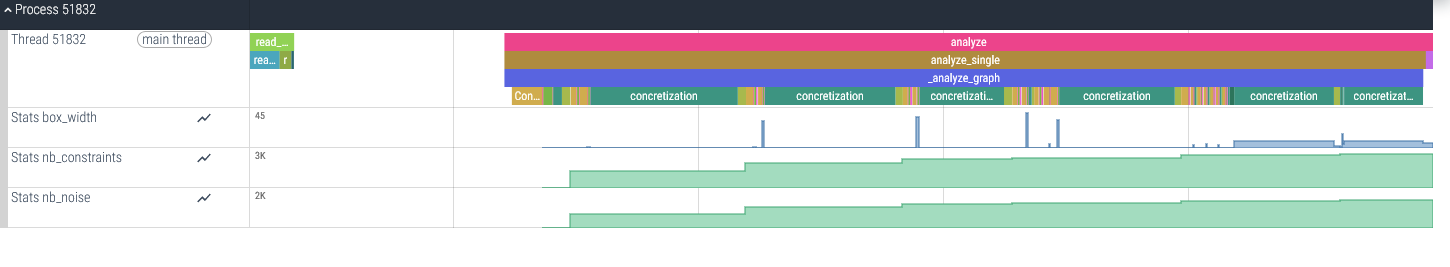}
    \caption{Profiling in \pyrat.}
    \label{fig:profile}
\end{figure}

Using the Trace Event Format and the \texttt{trace-events} library, \pyrat is also able to provide some profiling of the analysis (see Figure~\ref{fig:profile}). This profiling can include the time taken for the analysis of each layer, the current domain width, but also the number of constraints or noise symbols at any given point of the analysis. More information can be added manually in the profiling. This profiling can allow either a user to analyse where his analysis takes time or where precision is lost to improve \pyrat parameters or the network itself. It can also allow developers in \pyrat to optimise the analysis and its implemntation.

Following an analysis, \pyrat will provide four possible outputs: "True", the property is verified; "False", the property is false and a \cex is provided; "Unknown", it could not be verified nor falsified with the given options; "Timeout", the analysis stops when the (optional) timeout is reached.

\section{Experiments}

This section details some experiments to assess the different domains implemented in \pyrat, \ie their strengths and weakness on different benchmarks. These experiments were run on an Intel Core i9-11950H 2.60GHz CPU as well as an NVIDIA RTX A3000 GPU. For reproductibility, we used a single thread, even when using \bab.

Most of our experiments are performed on benchmarks used in the \vnncomp. For every benchmark in the \vnncomp, properties are randomly perturbed with an unknown seed. This explains the slight differences between our experiments and the one in the competition.

\subsection{Mooring line failure detection}

We first consider a small example with four safety properties on a single model. The model is a neural network with 7 inputs and 5 outputs performing a classification task. It consists of 3 hidden fully connected layers of 25 neurons each. This network was developed for mooring line failure detection on offshore platforms~\cite{technip} and has already been used in~\cite{reciph} to compare \bab on input heuristics in \pyrat. The safety properties considered here are functional properties on the network with large possible input variations.

Table~\ref{tab:toy-ex} shows the results of \pyrat's analysis with different domains and options. As a baseline, we indicate the time and the result of the analysis without any \bab for \zonos and \conzs. These analyses are inconclusive because the properties have a large input space and have from 60\% to 100\% of unstable \relu. With both \bab approaches, all properties are proved with similar time for \zonos with \bab on input and \conzs with \bab on \relu. Through Table~\ref{tab:nb-analysis}, we see the number of reachability analyses performed by \pyrat for all approaches. For input \bab, we see that the number of split required for \conz is significantly less than for \zono as we are more precise, but the overhead in computation time makes the overall analysis slower. The average analysis time on these networks is around 0.006s for \zono and 0.03s for \conz. 

\bab on \relu further reduces the number of analyses needed, which compensates for the time overhead. Our heuristics for choosing the best \relu to split on allow us to reduce the number of split by prioritizing the correct \relu nodes. Finally, \hybz are able to prove the four properties in only 6.28s with most of the time spent in the MILP solver (here we use the \texttt{cvxpy} library with the \texttt{Gurobi} solver). This is more than ten times faster than \bab approaches. As the network is small and the number of neurons is low, the MILP solver are able to solve this problem very quickly with our exact representation thus outperforming the other methods that need to over-approximate the network.

\begin{table}[ht]
\centering
\begin{tabular}{|l|c|c|c|}
\hline
 & \textbf{Unknown} & \textbf{Proven} & \textbf{Time} \\ \hline
\textbf{Zonotopes} & 4 & 0 & 0.13 \\ \hline
\textbf{Constrained zonotopes} & 4 & 0 & 0.19 \\ \hline
\textbf{Zonotopes w/ BaB input} & 0 & 4 & 94.88 \\ \hline
\textbf{Const. zono. w/ BaB input} & 0 & 4 & 140.49 \\ \hline
\textbf{Const. zono. w/ BaB ReLU}  & 0 & 4 & 95.25 \\ \hline
\textbf{Hybrid zonotopes} & 0 & 4 & 6.28 \\ \hline
\end{tabular}
\caption{Results of different analysis on the `Mooring line` example.}
\label{tab:toy-ex}
\end{table}

\begin{table}[ht]
\centering
\begin{tabular}{|c|c|c|c|}
\hline
\textbf{Property} & \textbf{Zono. \bnb input} & \textbf{Constr. zono. \bnb input} & \textbf{Constr. zono. \bnb \relu} \\ \hline
1 & 879 & 445 & 323\\ \hline
2 & 10113 & 2967 & 1067 \\ \hline
3 & 1369 & 607 & 253  \\ \hline
4 & 1763 & 557 & 657 \\ \hline
\end{tabular}
\caption{Number of analysis for \bab approaches on the same example.}
\label{tab:nb-analysis}
\end{table}

\subsection{LinearizeNN \& ACAS-Xu}

We then compare our domains on both the \texttt{LinearizeNN} and the \texttt{ACAS-Xu} benchmarks from the \vnncomp 2024 and we refer the reader to the competition report~\cite{vnn2024} for the full details of these benchmarks. Both benchmarks have a small number of inputs: 4 for \texttt{LinearizeNN} and 5 for \texttt{ACAS-Xu}. The models are fully connected models with varying depth and from 100 to 256 neurons on each hidden layer. We compare our abstract domains on the 60 and 186 safety properties of \texttt{LinearizeNN} and \texttt{ACAS-Xu}. The timeout for \texttt{LinearizeNN} is at 60 seconds and at 116 seconds for \texttt{ACAS-Xu}.

These benchmarks are more difficult to verify than the previous example, as shown in Table~\ref{tab:linearizenn} and Table~\ref{tab:acasxu}. The baseline with only \zono or \conz proves or falsifies (by finding a \cex) a small number of properties of the benchmark and we can see the increase in precision of \conz as compared to simple \zono. The tables clearly show a better performance for the methods based on \bab on input rather than \bab on \relu. In fact, the number of inputs is much smaller than the number of unstable \relu to split (up to 500 unstable \relu on \texttt{LinearizeNN}). While on \texttt{LinearizeNN} \zonos have a slight edge in speed, on \texttt{ACAS-Xu}, where the properties are longer to prove, \conzs are faster overall because their increased precision outweighs the overhead.
Additionally, we observe that \hybzs barely prove more properties than \conzs on \texttt{LinearizeNN} while performing even worse on \texttt{ACAS-Xu}. As the number of unstable \relu grows, the generated MILP problem becomes too large to solve within the given timeout.
The number of properties falsified in Table~\ref{tab:linearizenn} and \ref{tab:acasxu} varies in function of the domain used and the \bab approach. Indeed, the property can be falsified if the entire output set of the reachability analysis does not satisfy the property. Thus, the precision of the abstract domain used plays an important role. During \bab multiple analyses are done on smaller input sets leading to more precision and more search for counter examples. In turn, the number of falsified properties is higher with \bab approaches and may vary depending on the number of analyses done.

\begin{table}[ht]
\centering
\begin{tabular}{|l|c|c|c|c|}
\hline
\textbf{}                          & \textbf{Falsified} & \textbf{Proven} & \textbf{Unknown} & \textbf{Time} \\ \hline
\textbf{Zonotopes} & 0 & 3 & 57 & 11.67 \\ \hline
\textbf{Const. zono.} & 0 & 10 & 50 & 51.58 \\ \hline
\textbf{Zonotopes w/ BaB input} & 2 & 58 & 0 & 123.34 \\ \hline
\textbf{Const. zono. w/ BaB input} & 2 & 58 & 0 & 191.89 \\ \hline
\textbf{Const. zono. w/ BaB ReLU}  & 0 & 19 & 41 & 2520.74 \\ \hline
\textbf{Hybrid Zonotopes} & 0 & 11 & 49 & 3181.84 \\ \hline
\end{tabular}
\caption{Results of different analysis on the \texttt{LinearizeNN} benchmark.}
\label{tab:linearizenn}
\end{table}

\begin{table}[ht]
\centering
\begin{tabular}{|l|c|c|c|c|}
\hline
\textbf{}                          & \textbf{Falsified} & \textbf{Proven} & \textbf{Unknown} & \textbf{Time} \\ \hline
\textbf{Zonotopes} & 13 & 4 & 169 & 27.16 \\ \hline
\textbf{Const. zono.} & 13 & 42 & 131 & 137.13 \\ \hline
\textbf{Zonotopes w/ BaB input} & 45 & 133 & 8 & 3376.74 \\ \hline
\textbf{Const. zono. w/ BaB input} & 47 & 138 & 1 & 2921.28 \\ \hline
\textbf{Const. zono. w/ BaB ReLU}  & 42 & 83 & 61 & 11022.07 \\ \hline
\textbf{Hybrid Zonotopes} & 14 & 22 & 150 & 18184.11 \\ \hline
\end{tabular}
\caption{Results of different analysis on the \texttt{ACAS-Xu} benchmark.}
\label{tab:acasxu}
\end{table}

\subsection{Cifar100}

Finally, the \texttt{cifar100} benchmark (also from the \vnncomp) is used to illustrate \pyrat's domains on a larger model and with higher input dimensionality. We limit the experiment here to the medium size network of the benchmark with 2.5 million parameters including convolutional layers, residual connections and \relu activations. There are 100 safety properties to check, corresponding to a local robustness property on 100 images from the CIFAR-100 dataset (32x32 pixels and 100 output classes). The experiments here are run using GPU with a timeout of 100 seconds.

As seen in Table~\ref{tab:cifar100-res}, the \cex search using adversarial attacks finds 12 \cexs while the \zono analysis does not prove any property. Using \conz, we can directly prove 8 properties without \bab and 32 out of 100 with \bab on \relu. Due to the high dimensionality of the input, the \bab on inputs with \conzs does not prove any additional property. On this benchmark, \hybzs fail to scale and are therefore not included in these results as they timeout on all properties. Overall, we only prove less than half of the 100 properties of this benchmark; future work will aim to increase this number by improving the speed of analysis to perform more \bab but also improving \bab heuristics as \bab on \relu is highly dependant on the chosen \relu nodes.

\begin{table}[ht]
\centering
\begin{tabular}{|l|c|c|c|c|}
\hline
 & \textbf{Falsified} & \textbf{Proven} & \textbf{Unknown} & \textbf{Time} \\ \hline
\textbf{Zonotopes} & 12 & 0 & 88 & 100.05 \\ \hline
\textbf{Const. zono.} & 12 & 8 & 80 & 261.72 \\ \hline
\textbf{Const. zono. w/ \bnb input} & 12 & 8 & 80 & 8216.16 \\ \hline
\textbf{Const. zono. w/ \bnb ReLU} & 12 & 32 & 56 & 6277.12 \\ \hline
\end{tabular}
\caption{Results of different analysis on the \texttt{cifar100} benchmark.}
\label{tab:cifar100-res}
\end{table}

\section{Applications}
\paragraph{Functional properties}

The first application of \pyrat is the verification of functional properties on structured data. On top of the usual ACAS-Xu application and the properties defined by \tool{Reluplex} \cite{katz2017reluplex}, \pyrat was used in a collaboration with Technip Energies to verify artificial neural networks for mooring line failure detection \cite{technip} on which formal safety properties were defined and verified. These small applications mainly rely on \pyrat's input splitting technique due to their low dimensionality and large input space to prove.

\paragraph{Local robustness} 

Through several projects such as the Confiance.ai program\footnote{https://confiance.ai/}, \pyrat has been used in industrial use cases, including large image classification neural networks. These projects were an opportunity to confront formal methods with industrial use cases where inputs of 224x224 pixels can be considered small. To be able to provide local robustness guarantees with \pyrat on such use cases, improvements had to be made (use of GPU, reduction of RAM usage, ...). Additionally, training techniques using abstract interpretation techniques~\cite{mirman2020} were used to train models that are easier to verify on 224x224 pixel dataset. On these models, which achieved 95\% accuracy, \pyrat was able to prove more than 90\% of the test set to be locally robust to small perturbation~\cite{rob_train}. \pyrat is also being used in the TRUMPET project~\cite{trumpet} to verify local robustness properties on privacy-enhanced NN for medical applications.

\paragraph{Embedded AI \& quantized networks}
\pyrat was used to verify embedded NNs and their Operational Design Domain (ODD) on the ACAS-Xu use case in a collaboration with Airbus~\cite{gabreau2024}. Following the methodology described in~\cite{damour2021}, the verification was performed on quantized neural networks in int8. Although these networks use quantized operators such as QLinearMatmul from the ONNX library, these operators still rely in part on floating point numbers. In this sense, \pyrat had to implement a new abstraction for conversion from float to integer datatype. In fact, since most of the MatMul operator is done with integer, the network is less prone to floating point errors and \pyrat is able to be correct \wrt floating points by correctly representing this floating point error for the remaining float operations.
In addition, splitting mechanisms~\cite{reciph} were also used by \pyrat to improve the ODD verification. Through this work, Airbus was able to produce a hybrid system using neural networks when the ODD is verified by \pyrat and present a certification approach according to the ED-324/ARP6983 standard. 
While initially used in embedded environments, quantized neural networks are not limited to this as quantification approaches are becoming a mainstream practice to reduce the size and footprint of the models in all environments. The need for safety of quantized networks and the potential use of \pyrat is therefore not limited to embedded AI systems.

\paragraph{CAISAR~\cite{girardsatabin2022}} is an open-source platform for evaluating the trustworthiness of AI through a unified entry point and high-level specification language using WhyML. In addition to its standalone use, \pyrat is closely integrated with CAISAR in order to provide a wider range of properties beyond classical safety or local robustness.

\begin{figure}[h]
\centerline{\includegraphics[width=\textwidth]{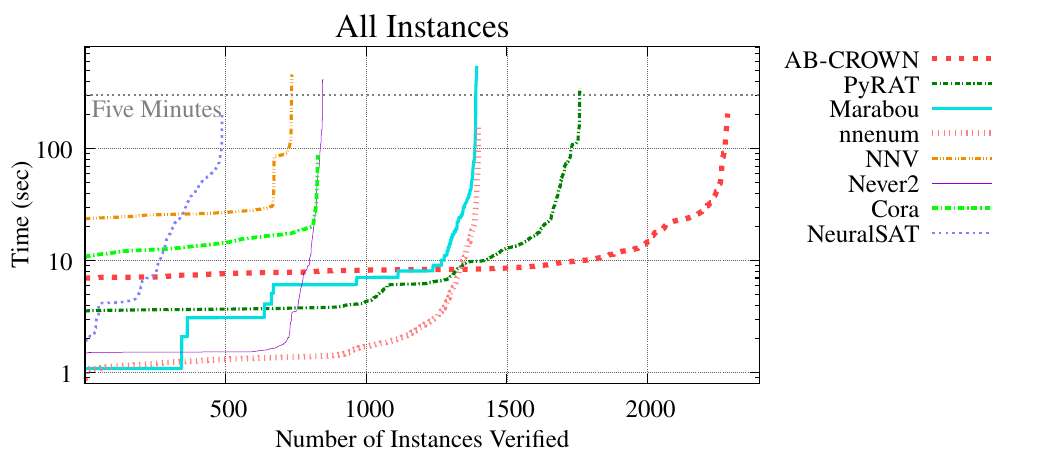}}
\caption{\vnncomp 2024 results on all instances (from \cite{vnn2024})}
\label{fig:res_vnn}
\end{figure}

\paragraph{\vnncomp} 
\pyrat participated in 2023 and 2024 in the international neural network verification competition on a wide variety of benchmarks \cite{vnn2023,vnn2024} reaching respectively the 3\textsuperscript{rd} and 2\textsuperscript{nd} place. From these participations, we can see the improvements made in \pyrat with new supported benchmarks, \eg \texttt{ml4acopf} or \texttt{cifar100}, and overall verification improvements.
Figure~\ref{fig:res_vnn} shows that \pyrat can verify small instances quickly, while for more complex instances is only second to \abcrown. 
Its wide support of layers and architecture allows \pyrat to prove more properties than others.\\
Yet, \pyrat has approximately 500 more timeout than \abcrown. 400 of them happen on "Safe" properties that \pyrat could not verified. Digging into the detailed results of the report, we found that \pyrat falls behind on larger input space problems such as \texttt{cifar100}. It highlights some shortcomings of \pyrat's \bab methods on \relu, and the need to improve and generalize the heuristics and \bab methods in \pyrat on non-linear activation functions. On the contrary, we found that \pyrat performs particularly well on problems with small input spaces which can be explained by the cheapness of the \zono domain combined with a well-calibrated \bab on input.

\paragraph{Soundness Benchmark~\cite{zhou2024}} is a benchmark to assess the correction of neural network verifiers. They propose a training procedure to hide counterexamples from verifiers and evaluate the state-of-the-art verifiers on different architecture. \pyrat is one of the two tools that are assessed as correct on this benchmark.

\section{Future works}
This paper presents the PyRAT tool, a neural network verification tool based on abstract interpretation with multiple domains such as zonotopes. Currently, \pyrat can verify the safety of a large range of neural networks of different sizes leveraging CPU and GPU as evidenced by its second place at the \vnncomp 2024. Current works in \pyrat already focus on generalising our approaches for non-linear functions as well as more precise abstraction. Efforts to verify architectures such as RNN and Transformers have also started and will be the subject of future works. With the development of edge AI and the proliferation of IoT devices, \pyrat will thrive to extend its support to embedded AI systems with their specific problems, \eg mixing integer and floating-point numbers, new rounding functions, etc. This work will be the subject of collaboration with the Aidge platform\footnote{https://eclipse.dev/aidge/}. Finally, \pyrat will also improve and extend its \bab techniques to new layers and with new improved heuristics to guide the \bab.

\paragraph{Acknowledgment}

This work has been supported by the French government under the ”France 2030” program, as part of the SystemX Technological Research Institute, and as part of the DeepGreen project with grant ANR-23-DEGR-0001. \pyrat has also been funded under the Horizon
Europe SPARTA project grant no. 830892 and TRUMPET project grant no. 101070038 as well as the European Defence Fund AINCEPTION project grant no. 101103385.

\paragraph{Main contributors} Augustin Lemesle, Julien Lehmann, Serge Durand (past), Tristan Le Gall, Samuel Akinwande (past).

\bibliographystyle{unsrtnat}
\bibliography{biblio}

\end{document}